# System Network Analytics:
# Evolution and Stable Rules of a State Series

Animesh Chaturvedi
*Data Science and Intelligent Systems*
*Indian Institute of Information Technology Dharwad*
Dharwad, Karnataka, India
animesh.chaturvedi88@gmail.com

Aruna Tiwari
*Computer Science and Engineering*
*Indian Institute of Technology Indore*
Indore, India
artiwari@iiti.ac.in

Nicolas Spyratos
*Computer Science Department - LISN Lab*
*University of Paris-Saclay*
Paris, France
spyratos@lri.fr

*Abstract*— **System Evolution Analytics** on a system that evolves is a challenge because it makes a *State Series* SS = {$S_1$, $S_2$... $S_N$} (i.e., a set of states ordered by time) with several inter-connected entities changing over time. We present *stability* characteristics of interesting *evolution rules* occurring in multiple states. We defined an *evolution rule* with its *stability* as the fraction of states in which the rule is interesting. Extensively, we defined *stable rule* as the evolution rule having stability that exceeds a given threshold *minimum stability* (minStab). We also defined *persistence metric*, a quantitative measure of persistent entity-connections. We explain this with an approach and algorithm for *System Network Analytics* (SysNet-Analytics), which uses minStab to retrieve Network Evolution Rules (NERs) and Stable NERs (SNERs). The retrieved information is used to calculate a proposed System Network Persistence (SNP) metric. This work is automated as a SysNet-Analytics Tool to demonstrate application on real world systems including: software system, natural-language system, retail market system, and IMDb system. We quantified stability and persistence of entity-connections in a system state series. This results in evolution information, which helps in system evolution analytics based on knowledge discovery and data mining.

I. INTRODUCTION

State of a system contains several distinct evolving inter-connected entities, which creates multiple networks referred as *evolving networks* [1] or *temporal networks* [2]. A set of evolving networks can be pre-processed to make a database series [3], which can also be referred to as dynamic databases or time variant data. The system state series is an unstructured (real-world) representation that needs pre-processing, whereas the database series is a structured representation to do data analytics. System state series are stored and managed in a repository (e.g., GitHub, Wikipedia, Maven, etc.) to keep the representations of continuous evolving states. A data analytics technique helps to understand and analyse database series.

A database of entities (or items) mined with *Association Rule Mining* [4] retrieves the association rules with *support* and *confidence*. The rule (X→Y) is formed with co-occurrence of entity sets: X (antecedent) and Y (consequent). The symbol → means co-occurrence such that if X occurs then Y will also occur. The sequence of entities in a database mined with *Sequential Rule Mining* (SRM) [5] retrieves *sequential rules* with support and confidence. The entities are unordered in association rules and ordered in sequential rules, prediction rules and temporal rules.

As the system evolves, its database also evolves, thus, its rules evolve as well. These rules help to study system evolution. Rule mining retrieves rules information of entities. Evolution and change mining [6] on a dataset over time retrieves the evolution information. We present network evolution rule mining to retrieve evolution rule information in the multiple states. This uncovers interesting and actionable information.

We detect stable link-patterns in evolving system networks over time. Some network states may have *persistent links* over time [7][8], thus their database states have *temporal association rules* over time [9]. We aim to retrieve *stable (or persistent) rules* from *evolution rules* with antecedent as source entities and consequent as target entities in a series of network databases. We have added characteristics of stability on links of evolving networks. The stability of links provides system evolution information in the network databases series to help *System Evolution Analytics* [10][11].

Given a state series SS = {$S_1$, $S_2$ ... $S_N$}, a rule might be interesting in one state, but not interesting in another state. Our approach identifies the evolution and stability information about inter-connected entities in state series. We contribute a kind of network evolution rule mining approach for source-target nodes in a set of evolving networks. Our approach mines two kinds of rules in a state series: Network Evolution Rules (NERs) and Stable Network Evolution Rules (SNERs). This paper is built upon our previous approach that focused mainly on minStab threshold and system changeability analysis to measure the changes over states [12].

We made three significant contributions: (a) we introduced Evolution and Stable rules; (b) we proposed System Network Databases, SNERs, and Persistence metric; (c) we presented a detailed algorithm named System Network Analytics (SysNet-Analytics). We demonstrate detailed experimentation with results and literature surveys on evolution and stability analysis.

Rest of the paper is presented as follows. In Section 2, we present novel definitions. Section 3 illustrates an example. In Section 4, we proposed our algorithms. Section 5 describes our prototype 'SysNet-Analytics Tool'. Section 6 presents the use of tools to do experiments. After these sections, related works are discussed in Section 7 and conclusions in Section 8.

## II. DEFINITIONS & CONCEPTS

We present formal definitions of the proposed approach. Let a set of ordered states as a state series $SS = \{S_1, S_2 \ldots S_N\}$ is given for an evolving system, where $S_i$ is an $i^{th}$ state. We start with the definitions of *Evolution rule* and *Stable rule* of the form $X \rightarrow Y$, where X is an antecedent set, and Y is a consequent set for the co-occurrence of the two sets (X and Y) of entities. Formal definitions are as follows.

**Definition 1:** Suppose a state series SS makes a database series $DS = \{D_1, D_2 \ldots D_N\}$, assume a rule ($X \rightarrow Y$) in database $D_i$ for state $S_i$. The rule is *interesting* in $D_i$, if its support and confidence exceeds given thresholds.

- A distinct rule occurring in multiple states is said to be an *Evolution Rule* that has some *stability* in the state series, where stability is the fraction of states in which the rule is interesting.
- An evolution rule is said to be a *Stable Rule* if its stability exceeds a given threshold named minimum stability (minStab).

Each state can make a dynamic database, which could be mined to generate association rules. The *interestingness* characterizes the association rules for a database. The *stability* characterizes the *evolution rules* and *stable rules* for a database series. The interestingness and stability are the probabilistic measure to form the rule ($X \rightarrow Y$). The stability is a new measure for characterizing evolution rules over time and constitutes one of the main conceptual contributions. The functional meaning of stability is the measure of evolution that happened in an evolution rule. Stable rules provide a set of persistently co-occurring entities over a database series.

Assume a system state series is pre-processed to make a set of evolving networks represented as $EN = \{EN_1, EN_2 \ldots EN_N\}$, such that the evolving network $EN_i$ represents state $S_i$. Each evolving network contains many connections between entities that make *connection pairs*. A connection pair (CP) is defined as (L, R), where the symbols L and R as mnemonics for Left and Right, respectively. A *connection pair* in $S_i$ is an ordered pair of two subsets: source entities L followed by their target entities R. A source entity e in source set L has a target entity e' in target set R. The e and e' form a directed connection from e to e' in the network of state $S_i$. A collection of *connection pairs* makes a dynamic database for Network Rule Mining.

**Definition 2:** A *system network database* (SysNetDb) is made-up of a set of *connection pairs* $\{CP_1, CP_2 \ldots CP_M\}$, where M is the total number of connection pairs. The SysNetDb of state $S_i$ is denoted as *SysNetDb_i*. Dynamic databases for a state series are *SysNetDbs* = {SysNetDb_1, SysNetDb_2 … SysNetDb_N}.

A connection pair (in SysNetDb) has two ordered entity-sets, which resembles it with sequence (in sequence database). The SysNetDb is a kind of sequence database, which converges Network Rule Mining (NRM) to the Sequential Rule Mining (SRM) that can generate sequential rules as network rules. The NRM is a special case of SRM because the input is a sequence database (SysNetDb) and the resulting sequential (or network) rules contain ordered sets of source and target entities. A network rule has a characteristic of interestingness, which measures support and confidence with respect to a given state. In a network rule, the *antecedent* X contains frequent source entities co-occurring with the *consequent* Y containing frequent target entities.

We define three kinds of rules: the Network Rule, the Network Evolution Rule (NER) and the Stable NER. All of them have the form $X \rightarrow Y$, but they have different characteristics. We shall say that a set $X \cup Y$ of entities occurs in a connection pair (L, R), if $X \cup Y$ is a subset of $L \cup R$. Assume X is a subset of source entities in L and Y is a subset of target entities in R.

Intuitively, Network rule or NER ($X \rightarrow Y$) in a state means, "presence of the source entities (X) implies presence of the target entities (Y) in a sufficiently high fraction of connection pairs (in a SysNetDb)". Intuitively, Stable NERs are "the rules of co-occurring persistent (stable) entities over a set of graphs". A NER and a SNER have an additional characteristic of stability, which is defined with respect to a given state series. The collection of network rules (from multiple states) makes NERs. A NER is referred as SNER in a state series, if it exceeds the expert specified three thresholds as minSupCont-minConf-minStab. The minSupCount and minConf characterizes network rules in a state $S_i$, and the minStab characterizes NERs over a state series SS.

**Definition 3:** Let a SysNetDbs contains a set of entities $X \cup Y$ to form a rule $X \rightarrow Y$ of a state series SS such that

- The *support* of $X \rightarrow Y$ in $S_i$ is denoted and defined by: $sup(X \rightarrow Y, S_i) = minSupCount \div M$, where minSupCount is the support count of connection pairs in which $X \cup Y$ occurs and M is the total number of connection pairs in SysNetDb.
- The *confidence* of $X \rightarrow Y$ in $S_i$ is denoted and defined by: $conf(X \rightarrow Y, S_i) = sup(X \cup Y, S_i) \div sup(X, S_i)$.
- A *network rule* ($X \rightarrow Y$) is called *interesting*, if its support (or support count) and confidence are greater than thresholds minSupCount and minimum confidence (minConf), respectively, provided by an expert.
- The $X \rightarrow Y$ is a *Network Evolution Rule* (NER), if it is a distinct network rule present in multiple states. Here, distinct means considering identical rules only once.
- The *stability* of NER ($X \rightarrow Y$) in state series (SS) is denoted and defined by $stab(X \rightarrow Y, SS) = stabilityCount \div N$, where stabilityCount is the number of states in which $X \rightarrow Y$ is interesting and N is the cardinality (number) of states in SS.
- A NER $X \rightarrow Y$ is *Stable NER* in SS if its stability count is greater than ($\geq$) a threshold *minimum stability* (minStab) = minStabCount $\div$ N.

Next central concept of our work is the *persistence metric*, which uses minStab, count of evolution rules, and count of stable rules. The intuitive significance of the persistence metric is to measure the persistence of entity-connections. In a state series SS, we can measure the persistence of entity-connections in an evolving system based on two obvious facts.

**Fact 1:** if there is a high minStab (fraction of minStabCount over the number of states), then the system has many persistent entity-connections (i.e., undergone few changes in the connections over states). Inversely, if there is a low minStab, then the system has few persistent entity-connections (i.e., undergone numerous changes in the connections over states). This implies *number of persistent connections is directly proportional to the minStab*

$$\text{Persistent Connections} \propto minStab = \frac{\text{minStabCount}}{N} \quad \ldots (1).$$

**Fact 2:** if there are many *stable rule*s, then it reflects many persistent entity-connections among states. Inversely, if there are few *stable rule*s, then it reflects few persistent entity-connections among states. This implies *number of persistent connections is directly proportional to the stable rule count and inversely proportional to the distinct evolution rule count*

$$\text{Persistent Connections} \propto \frac{\text{SR\_Count}}{\text{ER\_Count}} \quad \ldots (2).$$

where, SR_Count stands for the StableRule_Count and ER_Count stands for the EvolutionRule_Count. These two facts (i.e., eq. (1) and (2)) are combined to define the persistence metric that characterizes system evolution.

**Definition 4:** Given a state series $SS = \{S_1, S_2 \ldots S_N\}$ of an evolving system with its evolution and stable rules, then the *persistence metric* of the system is defined as

$$\text{Persistence metric} = minStab \times \frac{\text{SR\_Count}}{\text{ER\_Count}} \times 100 \quad \ldots (3).$$

Analogously, we express a formula in context of system network, named as *System Network Persistence metric* (SNP metric) and defined as

$$\text{SNP metric} = minStab \times \frac{\text{SNER\_Count}}{\text{NER\_Count}} \times 100 \quad \ldots (4)$$

where, SNER_Count stands for the SNER count and NER_Count stands for the NER count.

The persistence metric can be the inverse of metrics like changeability (or evolvability), but both are different to each other and indicate entirely different physical significance. Persistence indicates resistance to change, and changeability provides tendency towards change. Both of the quantities are related to the amount of evolution happening to a system over time. Two important remarks regarding the persistence metric:

1) On one hand, low values of minStab and SR_count produces low values of the persistence metric, which means that few stable rules occur in a smaller number of states. On the other hand, high values of minStab and SR_count produces a high value of *persistence metric*, which means that many stable rules occur in many states. Low value reflects few persistent entity-connections and high value reflects many persistent entity-connections. The lower bound of persistence metric is 0 for a volatile system (that has no common rule between any two states). Conversely, the upper bound of the persistence metric is 100 for a constant system (that does not change with time and has identical states). Both lower and upper bound are ideal conditions, which rarely occur with a normal system. In a real-world system, persistence metric value varies due to three reasons: the system domain, the fraction of (minStabCount ÷ N), and the fraction of (SR_count ÷ ER_count).

2) We can use the *persistence metric* to know about the stability of a new state. Suppose the *persistence metric* is $PM_N$ for N states. If we add a new state to the system, then the new *persistence metric* is $PM_{N+1}$ for N+1 states. Following three conditions are possible $PM_{N+1} > PM_N$ or $PM_{N+1} < PM_N$ or $PM_{N+1} = PM_N$. If $PM_{N+1} > PM_N$, then changes are significantly done to construct a new state. If $PM_{N+1} < PM_N$, then the new state has not changed much as compared to the old states. If $PM_{N+1} \cong PM_N$, then the new state is like the old states.

Next section describes the basics of pre-processing to make a set of connection pairs. The experiment section describes a mechanism to gather connection pairs using connection relationship (Table I, column 5) for making SysNetDbs.

## III. An Illustrative Example

This section discusses an illustration for pre-processing and Network Rule Mining (NRM) of a state, followed by NERs and Stable NERs retrieval for a state series.

**Pre-processing of a State**: Fig. 1 shows pre-processing of a state $S_i$ to make *SysNetDb_i_ID* to perform NRM that generates network rules. First, pre-process a state series to make a series of evolving networks, where a node represents an entity and an edge represents directed entity connection. The pre-processing of state $S_i$ depends upon the type of system i.e., different systems have different techniques to create evolving networks. For example, pre-processing on a software code generates call graph and data flow graph to perform inter-procedural call analysis and data flow analysis. Pre-processing of natural-language generates word-network to perform natural-language processing. Second, use the evolving networks to gather a set of connection pairs with respect to a *connection relationship* between entities. For example, the sequence of words in a sentence decides a connection pair of words (as entities). Thus, formation of SysNetDb depends upon the connection relationship chosen to form connection pairs. The connection relationship decides the connection pairs to make a set of SysNetDbs.

Assume $E = \{e_1, e_2, \ldots e_K\}$ be a set of entities, and *SysNetDb_i_ID* = $\{CP_1, CP_2 \ldots CP_M\}$ is a set of connection pairs. The value of K depends on the entities in the network and multiple network states have different numbers of entities. In Fig. 1, assume a connection relationship provides 3 connection pairs with 5 entities in the network of state $S_i$ then K=5 and M=3. For $S_i$, pre-processing makes *SysNetDb_i* with connection pairs (L, R) in the form of *entityName* that are indexed to *entityID*. The Indexing process is as follows. First, each entityName appearing in $S_i$ are enumerated with a unique positive integer as entityID. Second, for each connection pair (L, R) in *SysNetDb_i*, the entities in L and in R are replaced with their entityID. This Indexing makes *SysNetDb_i_ID*, which helps in faster, optimized, and efficient network rule mining.

**Network Rule Mining on a state** takes 3 inputs (SysNetDb, minSupCount, and minConf) to generate several network rules. Search space of NRM is SysNetDb of a state that contains connection pairs. The NRM has the following steps.

First, NRM scans the SysNetDb of state $S_i$ to calculate the frequency of each entity and then identifies all entities with frequencies greater than minSupCount.

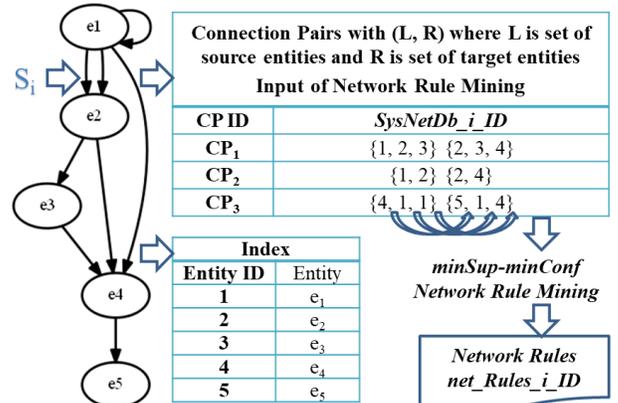

Fig. 1. An overview of pre-processing and network rule mining on SysNetDb_i_ID (contains three connection pairs) of a state $S_i$.

Second, NRM calculates the support (frequency) and confidence (conditional probability) for co-occurrence of antecedent X and consequent Y. Third, NRM computes the support and confidence of a network rule X→Y in a state $S_i$ according to the connection pairs in *SysNetDb_i_ID*. Finally, it uses the threshold minSup and minConf to generate ordered network rules X→Y in the form of ID (*net_Rules_i_ID*). Output of NRM is network rules with support and confidence greater than user-defined threshold minSupCount and minConf.

**Stable Network Evolution Rules of a State series:** For all the states in a state series, apply the NRM to retrieve network rules. Fig. 2 continues the example given in Fig. 1 by assuming that *SysNetDb_i_ID* as *SysNetDb_1_ID*. Here, the number of states (N) is three. In Fig. 2, the interesting network rules are generated according to minSupCount = 2 and minConf = 0.5. Thereafter, we merged the network rules of the three states to create a collection. From the collection, we identify distinct NERs with their stabilities. Then, we retrieved the SNERs from NERs whose Stability Count is greater than minStabCount = 2.

**System Network Persistence Calculation:** For illustration in Fig. 2, the SNP metric given in equation (4) uses minStabCount = 2, N = 3, SNERs = 5, and NERs = 11. The calculation will result in $\{(2 \div 3) \times (5 \div 11)\} \times 100 = 30.30$. Thus far, we presented novel definitions, concepts, and an illustrative example. Next, we present proposed algorithms.

## IV. SYSTEM NETWORK ANALYTICS: ALGORITHM

This section describes *System Network Analytics* (SysNet-Analytics), a theory to analyse stability of pre-evolved systems. Search space of the algorithm is a set of *evolving networks* EN = {$EN_1$, $EN_2$… $EN_N$} for a *state series* SS = {$S_1$, $S_2$ … $S_N$} at time points {$t_1$, $t_2$, $t_3$… $t_N$}. The *system network database* series is SysNetDbs = {SysNetDb_1, SysNetDb_2… SysNetDb_N}, whose $i^{th}$ database is represented as SysNetDb_i for a state $S_i$. We are interested to find SNERs having "the presence of source entities (X) implies the presence of target entities (Y) in a sufficiently high fraction of states". Our algorithm has the following four steps (also manifested in Fig. 3).

1. Pre-process N states stored in a repository to make N *SysNetDbs* and an *Index* with <*entityName, entityID*>.
2. Mining NERs and SNERs use four inputs (SysNetDbs, minSupCount, minConf, and minStabCount) to retrieve the *NERs_ID* and *SNERs_ID*.
3. The *NERs_ID* and *SNERs_ID* are in the form of *entityIDs*. Use the *Index* to replace each *entityID* with its *entityName* in each NER and SNER. This converts *NERs_ID* into *NERs_Name* and *SNERs_ID* into *SNERs_Name*.
4. Calculate SNP metric given in Equation (4) using the minStab, number of SNERs retrieved (SNER_Count), and number of NERs retrieved (NER_Count).

---
**Algorithm SysNet-Analytics(***repository***)**

Retrieve N system states and store them in a *repository*.
1. *SysNetDbs* & *IndexFile* = **Pre-process**(*repository*)
2. *NERs_ID* & *SNERs_ID* = **Mining_NERs_SNERs** (*SysNetDbs*, *minSupCount, minConf, minStabCount*)
3. *NERs_Name* & *SNERs_Name* = **Indexing**(*NERs_ID*, *SNERs_ID*, *IndexFile*)
4. *SNP_metric* = **SNP_Metric**(*minStab*, *SNER_Count*, *NER_Count*)

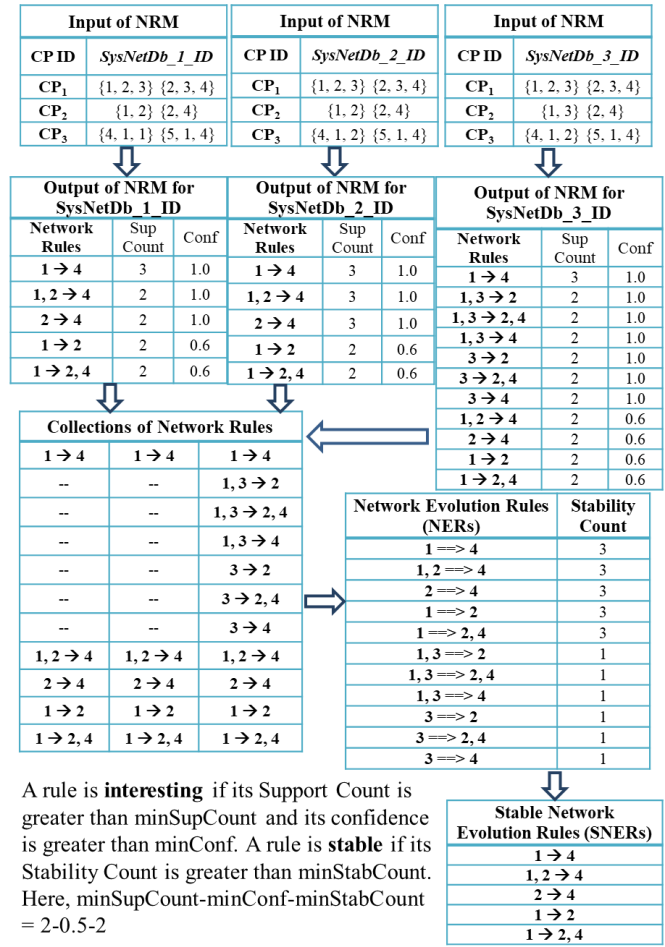

A rule is **interesting** if its Support Count is greater than minSupCount and its confidence is greater than minConf. A rule is **stable** if its Stability Count is greater than minStabCount. Here, minSupCount-minConf-minStabCount = 2-0.5-2

Fig. 2. The collection of network rules to make Network Evolution Rules (NERs) from which Stable NERs (SNERs) are retrieved.

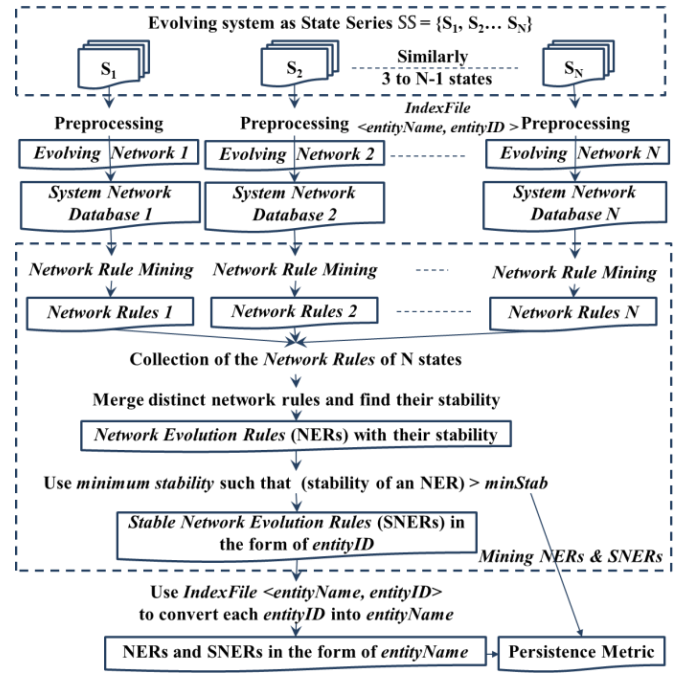

Fig. 3. SysNet-Analytics theory for a state series of an evolving system.

## Algorithm 1 Pre-process(*repository*)

Initialize HashMap *Index*< *entityName, entityID* >
Initialize integer *counter* = 1
Initialize String Buffer *buffer*

**For each** state $S_i$ where i ∈ integer and i varies from 1 to N
  Depending on type of repository, extract relationship between set of inter-connected entities to create a *SysNetDb_i* for a state $S_i$
  Read *SysNetDb_i* and store it in *buffer* until **end of file**
  **For each** line of *buffer*, **scan** *entityName*
    **If** an *entityName* is in the *Index*
      In *buffer*, replace the *entityName* with its *entityID*
    **Else**
      *entityID* = *counter*
      Add the new tuple < *entityName, entityID* > in the *Index*
      In *buffer*, replace the *entityName* with its *entityID*
      Increment the *counter* by 1 i.e. *counter* = *counter* + 1
  **End of Scan** when end of *buffer* is reached
  Write the *buffer* in *SysNetDb_i_ID*
  Store the file *SysNetDb_i_ID* in directory *SysNetDbs*
**End For** when all the states are pre-processed

Make an *IndexFile* and store *Index*<*entityName, entityID*>.

**Return** SysNetDbs & IndexFile

## Algorithm 2 Mining_NERs_SNERs(*SysNetDbs, minSupCount, minConf, minStabCount*)

Initialize File *net_Rules_i_ID NERs_ID, SNERs_ID*
Initialize Array *Collect_NRs_ID*,
Initialize HashMap *NERs_HM* < NER_ID, stability >
Initialize i ∈ integer

**For each** *SysNetDb_i_ID* in *SysNetDbs*, where i varies from 1 to N
  *net_Rules_i_ID*= **NRM**(*SysNetDb_i_ID, minSupCount, minConf*)
  Store *net_Rules_i_ID* file in directory *netRules*
**End For**

**For each** state *net_Rules_i_ID* in *netRules*, where i varies 1 to N
  *Collect_NRs_ID* = **Merge**(*Collect_NRs_ID, net_Rules_i_ID*)
**End For**

**For each** distinct rule (as *NER_ID*) in *Collect_NRs_ID*
  Initialize int *stabilityCount* = 0
  **For each** *rule_x* in *Collect_NRs_ID*
    **if**(*NER_ID* equal to *rule_x*)
      **then** *stabilityCount++*
    **end if**
  **End for**
  Initialize float *stability* = *stabilityCount* ÷ N
  Initialize float *minStab* = *minStabCount* ÷ N
  **if**(*NER_ID* is not in *NERs_HM*)
    **then Add**(<*NER_ID, stability*> to *NERs_HM*)
  **end if**
  **if**(*stability* > *minStab*)
    **if**(*NER_ID* is not in *SNERs_ID*)
      **then Add**(*NER_ID* to *SNERs_ID*)
    **end if**
  **end if**
**End For**

*NERs_ID* = *NERs_HM*
**Return** *NERs_ID* and *SNERs_ID*

### A. Pre-processing of a state series

The Algorithm 1 Pre-process takes N states to create N directed evolving networks that further makes N SysNetDbs for a connection relationship. A *SysNetDb_i* contains connection pairs in the form of *entityName*. Then, transform each *SysNetDb_i* to make *SysNetDb_i_ID* in the form of *entityIDs*. An entity occurring in several states must have the same *entityID*. Thus, an *entityID* keeps track of multiple occurrences of an entity in a state series. Simultaneously, an *Index* is also created, which has a list with two entries <*entityName, entityID*> for each entity. Output of pre-processing is N SysNetDbs & an *IndexFile*. The N SysNetDbs is a set of dynamic databases such that each SysNetDb follows the semantics, syntax, and structure for input required in NRM, which automates knowledge discovery. For reproducibility, we present pseudo code for pre-processing of a state series.

### B. Mining NERs and SNERs

The Algorithm 2 Mining_NERs_SNERs takes four inputs: SysNetDbs, minSupCount, minConf, and minStab. The minStab measures stability of a NER over a state series. The algorithm executes the NRM to process SysNetDbs and generates network rules for N states. Each *SysNetDb_i_ID* is processed to make network rules (*net_Rules_i_ID*) for state $S_i$. The network rules for N states are stored in N files that are collectively stored in a directory referred as *netRules*.

First, the algorithm merges network rules of N states to make a collection of rules as *Collect_NRs_ID*. This collection may have some similar rules and some dissimilar rules. Second, count the occurrence of each identical rule *NER_ID* in *Collect_NRs_ID*. The count represents *stabilityCount* of a *NER_ID*, which means the number of states in which *NER_ID* is interesting. Calculate *stability* as the fraction of *stabilityCount* and number of states (N). Then, add the *NER_ID* and its *stability* to the hash-map (*NERs_HM*). Make a file *NERs_ID* to store the retrieved *NERs* and their stability. Third, select a *NER_ID* as *SNER_ID*, if the *NER_ID* has more stability than minStab. Make a file SNERs_ID and store each SNER (X→Y) only once.

The threshold (minSupCount-minConf-minStabCount) is used to control the quality of output rules. A set of low values of threshold can generate an exhaustive number of rules, which may be useless in decision-making. A best possible high value of threshold optimizes the number of interesting NERs and SNERs. We determine the best possible high values of threshold based on well-known theory of exploration and exploitation. The threshold values are dependent on domain and SysNetDbs. Search minSupCount-minConf values, which generate the optimal number of NERs. For optimum low values of minSupCount-minConf, *explore NERs* by varying the value of minStabCount. After getting the optimum number of NERs, start searching for a range of minStabCount. In that range, start *exploiting SNERs* by varying the value of *minStab*. This results in the best possible high values of the three thresholds. The number of SNERs are optimized by using distinct NERs in each state. To help decision-making, each SNER (X→Y) represents a persistent link of co-occurring entities over network states.

## C. Indexing

The retrieved NERs and SNERs are in the form of *entityIDs*, which helps in optimized and faster mining. These *entityIDs* are hard to comprehend because it is in the number format. Thus, for better comprehension the Algorithm 3 Indexing replaces all the *entityIDs* in the *NERs_ID* and *SNERs_ID* with their corresponding *entityNames*. The algorithm uses the *IndexFile* produced in pre-processing to make two files *NERs_Name* and *SNERs_Name* that store rules in the form of *entityNames*. For example, the format of SNERs (X→Y) in *SNERs_ID* contains *entityID* like {1, 2}→{3}, which are similar to association rules with X = {1, 2} and Y = {3}. These rules can be converted to *entityName* format rules like {butter, jam}→{milk} where 1, 2, and 3 stands for butter, jam, and milk, respectively.

| Algorithm 3 **Indexing**(*NERs_ID*, *SNERs_ID*, *IndexFile*) |
|---|
| Initialize HashMap *Index*<*entityName*, *entityID*> |
| Initialize integer *counter* = 1 |
| Initialize String Buffer *buffer_1*, *buffer_2* |
| Make two files *NERs_Name*, *SNERs_Name* |
| **For each** line of *IndexFile* |
|     **Scan** line and **Store** <*entityName*, *entityID*> in *Index* |
| **End For** when *IndexFile* is completely scanned |
| **For each** line of file *NERs_ID*, |
|     **Scan** and **store** the line in *buffer* |
|     In *buffer*, **replace** each *entityID* with its *entityName* in *Index* |
|     **Store** the buffer in *NERs_Name* |
| **End For** when *NERs_ID* is completely scanned |
| **For each** line of file *SNERs_ID*, |
|     **Scan** and **store** the line in *buffer* |
|     In *buffer*, **replace** each *entityID* with its *entityName* in *Index* |
|     **Store** the buffer in *SNERs_Name* |
| **End For** when *SNERs_ID* is completely scanned |
| **Return** *NERs_Name*, *SNERs_Name* |

## D. SNP Metric

After retrieving the NERs and SNERs, finding their counts is a straightforward process. Keep note of the minStab used to retrieve SNERs. The Definition 4 SNP metric given in equation (4) uses: minStab, NER count, and SNER count. It measures the persistence quantitatively for entity-connections. The calculation of the SNP metric is self-explanatory.

## V. SYSNET-ANALYTICS TOOL

Based on Algorithm SysNet-Analytics, we developed an automated SysNet-Analytics Tool using Java technology (JRE and JDK). As input, the tool takes N *system network databases* (SysNetDbs) with thresholds. As output, the tool retrieves network rules, collection of NERs, and SNERs.

The tool discovers these system evolution rules using three components based on the three algorithms: '*Pre-processing*', '*Mining_NERs_SNERs*' and '*Indexing*'. First, pre-process a state series to make SysNetDbs. Second component of the tool uses network rule mining (NRM) on SysNetDb, which contains sequences of source and target entities. Our tool mines network rules common to several *connection pairs*. The tool extends the RuleGrowth algorithm [13]-[14] for NRM because a *connection pair* is used as a *sequence* to retrieve network rules. The tool merges unique network rules to make a collection of NERs. Then, find NERs with high stability to retrieve non-redundant SNERs. Third component converts the entity ID based rules to entity-name based rules as natural-language is more comprehensive. The tool summarizes the SNERs into a report file, which helps to predict missing and possible co-occurring interconnected entities.

This tool is interesting for practitioners, who deal with evolving networks representing connections between entities. The SysNet-Analytics retrieves stable and evolution rules as evolution information, which can be used to analyse the persistent connections over a time-period. A practitioner can apply similar steps of SysNet-Analytics to gain insightful information, which is useful due to evolution and stable characteristics of rule for inter-connected system entities.

The complexity of pre-processing a system depends on the system domain. This means every system domain has different complexity of preprocessing, thus assuming it as X. The algorithm Mining NERs_SNERs uses the NRM algorithm for N times on N states. Rest of the algorithm has $n^2$ complexity. Hence, the algorithm complexity mainly depends on the complexity of the NRM algorithm. Therefore, the overall complexity of the SysNet-Analytics is $X + N \times O(NRM) + n^2$. The algorithm is efficient because it is sequential in nature. However, it depends on the efficiency of the NRM algorithm, the computing resources for scalability, and the network size.

## VI. SYSTEM EVOLUTION ANALYTICS EXPERIMENTS

This section describes application of SysNet-Analytics algorithm using SysNet-Analytics Tool on six real-world evolving systems collected from open-internet repositories. This demonstrates how to automate and apply the SysNet-Analytics on different kinds of evolving systems, which proves the usefulness of automation achieved with our tool.

Pre-processing a state series is an expert domain task. With domain specific algorithms, we pre-processed each evolving system to generate its set of *evolving networks*. Each network is a directed graph with relationship information about inter-connected entities of a system. Then, we transformed a set of evolving networks to produce a series of SysNetDbs. While pre-processing, we make two assumptions: (a) the system contains several entities and (b) each entity might be calling zero, one, or more other entities. If entity e is directed to e' then the e is the source-entity and the e' is the target-entity of the connection. Each line of a network file contains two entities (as nodes) to represent a directed connection (as edge) from the source entity to the target entity.

In Table I, the first two columns describe the six evolving systems belonging to four different domains. Third column provides the number of states used in the experimentation. Fourth column describes the kind of source and target entities in an evolving system. Fifth column describes the 'type of relationship' between the entities, which depends on the domain of the evolving system. Sixth column provides a 'type of network' for an evolving system. Few networks have an existing well-known name (like *call graph* for software) and few networks do not have a name, thus, we name them like word network, purchase network etc. Seventh column contains the total number of entities used to make networks. Eighth column contains the average number of neighbours (entities). Ninth column presents calculation for the number of aggregated

TABLE I DOMAIN INFORMATION OF 6 EVOLVING SYSTEMS AND THEIR EVOLVING NETWORKS TO PERFORM SYSTEM EVOLUTION ANALYTICS.

| Domains of System Evolution Analytics | Evolving Systems | N | 'Source' and 'Target' Entities | Type of Relationship | Type of network | Number of entities | Average number of neighbours | Number of aggregated connections |
|---|---|---|---|---|---|---|---|---|
| (A) Software Evolution Analytics | Hadoop HDFS[1] | 15 | 'Caller' and 'Callee' Procedures | Procedural calls | Call graph | 3129 | 2.166 | 15×3129×2.166 = 101661.21 |
| (B) Natural-language Evolution Analytics | List of Bible Translation[2] | 13 | Words in 'Source biblical language' and 'English variant' | Translations | Word Network | 246 | 1.456 | 13×246×1.456 = 4656.288 |
| | List of Multi-sport Events[3] | 13 | Words in 'Titles (name)' and Scopes (region) of events | Regional names | Word network | 141 | 1.786 | 13×141×1.786 = 3273.73 |
| (C) Market Evolution Analytics | Retail Market[4] | 13 | 'Products description' and 'Customer IDs' | Purchases | Purchase network | 1872 | 7.204 | 13×1872×7.20 = 175219.2 |
| (D) Movie Evolution Analytics on Evolving IMDb System[5] | Positive sentiment[6] of movie genres[5] | 16 | 'Positive words in names' and 'genres' of movies | Sentiments | Positive sentiment network | 284 | 2.661 | 16×284×2.661 = 12091.58 |
| | Negative sentiment[6] of movie genres[5] | 16 | 'Negative words in names' and 'genres' of movies | Sentiments | Negative sentiment network | 510 | 3.303 | 16×510×3.303 = 26952.48 |

1. https://mvnrepository.com/artifact/org.apache.hadoop/hadoop-hdfs
2. https://en.wikipedia.org/wiki/List_of_English_Bible_translations
3. https://en.wikipedia.org/wiki/List_of_multi-sport_events
4. https://archive.ics.uci.edu/ml/datasets/Online+Retail
5. http://www.imdb.com/interfaces/
6. https://www.cs.uic.edu/~liub/FBS/sentiment-analysis.html

connections using: number of states, number of entities, and average number of neighbours. Different repositories contain different kinds of evolving systems, e.g., Maven contains Hadoop-HDFS library jars as data, Wikipedia contains natural-language as data, UCI repository contains retail market data, and IMDb (Internet Movie Database) contains movie genre data.

The different values of thresholds (minSupCount, minConf, and minStabCount) generate different numbers of rules (with varying interestingness and stability) as output. Low values of thresholds resulted in many NERs, which include less interesting and less stable rules. For low thresholds, such an exhaustive number of rules is tedious to manage. To find the optimum number of SNERs, we *explored* (minSupCount and minConf) and *exploited* (minStab) for the best possible high values of thresholds. This generates an optimized number of Stable NERs, helps in decision-making and taking actions.

A. Experiments of SysNet-Analytics on four domains

**1. Software Evolution Analytics for Call graph:** We pre-processed 15 jars of Hadoop-HDFS to make 15 *evolving call graphs* (networks) that further makes 15 SysNetDbs for 15 versions (states). The connections are inter-procedural calls in jars. The connection pairs are sets of source caller procedures and set of target callee-procedures. A retrieved SNER (create → convert) suggests that procedure 'create' is a caller procedure and 'convert' is a callee procedure. The rule suggests that the procedure 'create' frequently calls procedure 'convert'. This reveals to us that the changes in procedure 'convert' may also affect the procedure 'create'.

**2. Natural-language Evolution Analytics for two lists:** We used natural-language text from two Wikipedia pages. The connections between word networks are selected from the words occurring in the same lines. The connection pairs are sets of pre-occurring words followed by sets of post-occurring words.

First Wikipedia page dataset contains a *list of bible translations*, which mention the translations done from the 7th century until 2014, from source biblical languages (like Hebrew and Greek) to English variant languages (like Modern and Old English). We combined three tables: *incomplete*, *partial*, and *complete Bibles*. We made evolving networks using inter-connected words in columns of 'English variant' and 'Source' of bible such that each network is for a century. We extracted 13 evolving networks that were converted to 13 SysNetDbs for 13 centuries. Two retrieved SNERs (English, Modern → Greek) and (English → Vulgate) suggest that bible in the Modern English language is likely to be translated from Vulgate or Greek. Our tool automatically retrieved many other such evolution rules (NERs and SNERs).

Second dataset contains a *list of multi-sport events*, which mention events that happened since 1890's until 2015. We pre-processed this dataset to create evolving networks from connections between words in event's 'title' and 'scope' (regional, international, and provinces). Such that each network is for a decade. We extracted 13 evolving networks that were converted to 13 SysNetDbs for 13 decades. Two retrieved SNERs (Games, Asian → Regional) and (Games, World → International) suggest that the 'Asian' 'Games' are of type 'Regional' sports. Similarly, we can deduce 'World' 'Games' are of type 'International' sports. Our tool automatically retrieved many other such NERs and SNERs.

**3. Market Evolution Analytics for Retail-market:** We used online retail market data [15], which contains purchasing information between Dec 2010 and Dec 2011. To make evolving networks, we used *product descriptions* and *customer IDs* for each month in the following way. A word in product description is used as a source entity only if the word appears in more than or equal to 10 months. Similarly, a customer ID is used as a target entity only if the customer did purchase in more than or equal to 10 months. These frequent product words and customer IDs are used to create 13 evolving networks, which are further converted to 13 SysNetDbs for 13 months. The connections between purchase networks are selected based on the words in the purchase (product words) and the customer IDs. The connection pairs provide sets of product words followed by sets of customer IDs. A retrieved SNER "SUKI SHOULDER BAG → 17841" suggests that the product with description (source words) 'SUKI SHOULDER BAG' is frequently purchased by the customer (target ID) '17841'. For low threshold values, there are several SNERs, which are useful to do target marketing.

TABLE II SNP METRIC CALCULATIONS FOR NINE EXPERIMENTS ON EACH OF THE SIX EVOLVING SYSTEMS.

| | minSupCount-minConf-minStabCount | N | minStab | SNER Count | Total NER Count | NERs fraction | SNP metric | minSupCount-minConf-minStabCount | N | minStab | SNER Count | Total NER Count | NERs fraction | SNP metric |
|---|---|---|---|---|---|---|---|---|---|---|---|---|---|---|
| | Hadoop-HDFS as an Evolving Software System [16] | | | | | | | List of Bible translations as an Evolving Natural-language System | | | | | | |
| 1 | 4-0.6-5 | 15 | 0.33 | 2 | 2 | 1 | 33.33 | 3-0.2-2 | 13 | 0.15 | 25 | 3715 | 0.007 | 0.1 |
| 2 | 4-0.6-4 | 15 | 0.27 | 2 | 2 | 1 | 26.67 | 2-0.3-2 | 13 | 0.15 | 28 | 5397 | 0.005 | 0.08 |
| 3 | 4-0.4-5 | 15 | 0.33 | 3 | 5 | 0.6 | 20 | 3-0.3-2 | 13 | 0.15 | 16 | 3715 | 0.004 | 0.07 |
| 4 | 4-0.6-3 | 15 | 0.2 | 2 | 2 | 1 | 20 | 2-0.2-4 | 13 | 0.31 | 12 | 5644 | 0.002 | 0.07 |
| 5 | 4-0.2-7 | 15 | 0.47 | 2 | 5 | 0.4 | 18.67 | 3-0.3-3 | 13 | 0.23 | 6 | 3715 | 0.002 | 0.04 |
| 6 | 4-0.8-5 | 15 | 0.33 | 1 | 2 | 0.5 | 16.67 | 2-0.3-3 | 13 | 0.23 | 6 | 5397 | 0.001 | 0.03 |
| 7 | 4-0.4-3 | 15 | 0.2 | 4 | 5 | 0.8 | 16 | 2-0.3-4 | 13 | 0.31 | 3 | 5397 | 0.001 | 0.02 |
| 8 | 4-0.4-4 | 15 | 0.27 | 3 | 5 | 0.6 | 16 | 3-0.3-4 | 13 | 0.31 | 3 | 3715 | 0.001 | 0.02 |
| 9 | 4-0.8-6 | 15 | 0.4 | 0 | 2 | 0 | 0 | 2-0.2-5 | 13 | 0.38 | 0 | 5644 | 0 | 0 |
| | List of Multi-sport events as an Evolving Natural-language System | | | | | | | Evolving Retail Market System | | | | | | |
| 1 | 3-0.8-2 | 13 | 0.15 | 5 | 8 | 0.625 | 9.62 | 4-0.6-2 | 13 | 0.15 | 25 | 131 | 0.191 | 2.94 |
| 2 | 3-0.2-2 | 13 | 0.15 | 6 | 11 | 0.545 | 8.39 | 5-0.6-2 | 13 | 0.15 | 12 | 71 | 0.169 | 2.6 |
| 3 | 2-0.3-2 | 13 | 0.15 | 14 | 41 | 0.341 | 5.25 | 4-0.8-2 | 13 | 0.15 | 14 | 86 | 0.163 | 2.5 |
| 4 | 2-0.8-2 | 13 | 0.15 | 11 | 37 | 0.297 | 4.57 | 4-0.6-3 | 13 | 0.23 | 12 | 131 | 0.092 | 2.11 |
| 5 | 2-0.3-4 | 13 | 0.31 | 5 | 41 | 0.122 | 3.75 | 4-0.6-4 | 13 | 0.31 | 7 | 131 | 0.053 | 1.64 |
| 6 | 2-0.3-3 | 13 | 0.23 | 6 | 41 | 0.146 | 3.38 | 5-0.6-3 | 13 | 0.23 | 5 | 71 | 0.07 | 1.63 |
| 7 | 2-0.8-3 | 13 | 0.23 | 4 | 37 | 0.108 | 2.49 | 4-0.8-3 | 13 | 0.23 | 6 | 86 | 0.07 | 1.61 |
| 8 | 2-0.8-4 | 13 | 0.31 | 3 | 37 | 0.081 | 2.49 | 5-0.6-4 | 13 | 0.31 | 3 | 71 | 0.042 | 1.3 |
| 9 | 3-0.8-4 | 13 | 0.31 | 0 | 8 | 0 | 0 | 4-0.8-4 | 13 | 0.31 | 3 | 86 | 0.035 | 1.07 |
| | Positive sentiment of movie genres in Evolving IMDb System | | | | | | | Negative sentiment of movie genres in Evolving IMDb System | | | | | | |
| 1 | 2-0.3-3 | 16 | 0.19 | 21 | 146 | 0.144 | 2.70 | 2-0.3-3 | 16 | 0.19 | 31 | 258 | 0.12 | 2.25 |
| 2 | 2-0.3-4 | 16 | 0.25 | 5 | 146 | 0.034 | 0.86 | 4-0.3-3 | 16 | 0.19 | 7 | 79 | 0.089 | 1.66 |
| 3 | 2-0.3-5 | 16 | 0.31 | 4 | 146 | 0.027 | 0.86 | 2-0.3-4 | 16 | 0.25 | 10 | 258 | 0.039 | 0.97 |
| 4 | 3-0.3-4 | 16 | 0.25 | 4 | 74 | 0.054 | 1.35 | 4-0.3-4 | 16 | 0.25 | 3 | 79 | 0.038 | 0.95 |
| 5 | 4-0.3-3 | 16 | 0.19 | 4 | 49 | 0.082 | 1.53 | 3-0.3-4 | 16 | 0.25 | 4 | 116 | 0.034 | 0.86 |
| 6 | 3-0.4-4 | 16 | 0.25 | 3 | 60 | 0.050 | 1.25 | 3-0.4-4 | 16 | 0.25 | 2 | 84 | 0.024 | 0.6 |
| 7 | 3-0.5-4 | 16 | 0.25 | 2 | 48 | 0.042 | 1.04 | 3-0.5-4 | 16 | 0.25 | 1 | 63 | 0.016 | 0.4 |
| 8 | 4-0.3-4 | 16 | 0.25 | 1 | 49 | 0.020 | 0.51 | 4-0.3-5 | 16 | 0.31 | 1 | 79 | 0.013 | 0.4 |
| 9 | 4-0.3-5 | 16 | 0.31 | 0 | 49 | 0.000 | 0.00 | 2-0.3-5 | 16 | 0.31 | 3 | 258 | 0.012 | 0.36 |

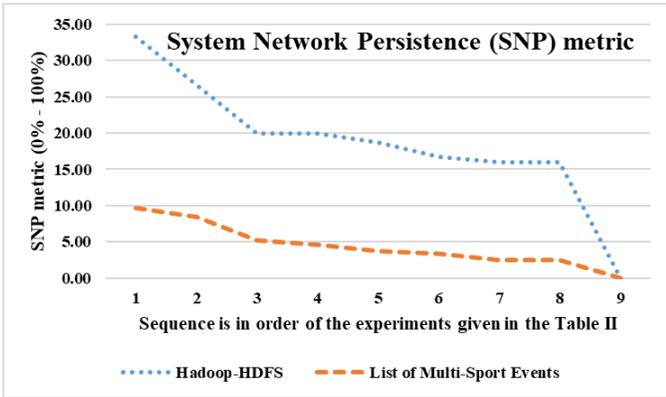
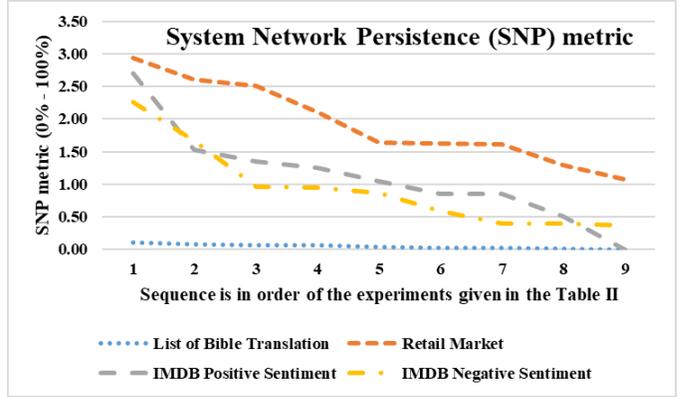

Fig. 4. SNP metrics using the nine experiments for all six evolving systems.

**4. Movie Evolution Analytics for IMDb:** We used movie name and genre data of IMDB[5] for 16 decades (1870's - Oct 2016). We made two types of evolving networks using two sentiment-list[6] of positive and negative words; Minqing and Bing [17]. Each connection in each network has the sentiment words in movie names as source entities and the genres as target entities. The connection pairs are sets of sentimental words followed by sets of genres. First type of evolving networks has connections between positive words in 'movie names' as source entities and their 'genres' as target entities. A retrieved SNER "premier → Short" suggests that the movie name containing 'premier' appears mostly in 'Short' genre movies. While naming a movie for a genre, SNERs are useful to find positive words, which are suitable for a genre and to target a positive audience. Second type of evolving network has connections between negative words in 'movie names' as source entities and their 'genres' as target entities. A retrieved SNER "sin → Drama" suggests that the movie name containing 'sin' appears mostly in the 'Drama' genre. SNERs are useful to find negative words that are suitable for a genre and to target a negative audience.

*B. System Network Persistence (SNP) Experimental Results*

The Table II has six parts, where each part reports the nine experiments for an evolving system, and each row represents an experiment. Each part has seven columns. First column mentions the experiment number (1 to 9) and the values of thresholds to do the experiment. Second column contains the number of states (N). Third column contains the fraction of minimum Stability {minStabCount ÷ N}, which provides stability of NERs over states. A NER with less minStab fraction is less stable, whereas a NER with high minStab fraction is more stable. Fourth column contains the number of SNERs retrieved in the experiment. Fifth column contains the distinct NER count retrieved in the experiment. Sixth column contains the fraction of stable NERs {(SNER count) ÷ (total NER count)}. Seventh column contains the value of the SNP metric for the experiment. Each part is sorted in increasing order of SNP metrics.

For all nine experiments of an evolving system, we present details about SNP metric (equation (4)). Fig. 4 has two plots with six time series to demonstrate persistence metric values for all six evolving systems. Each time series has nine coordinates that represent nine SNP metric values for the nine experiments. The X-axis in the figure has the same sequence as mentioned in Table II for each evolving system. For the time series of Hadoop-HDFS, in the X-axis the '1' represents the first experiment '1' mentioned in the first column of Table II for Hadoop-HDFS with 4-0.6-5.

Fig. 4 provides to measure persistence of entity-connections (0 to 100 in Y-axis) for the evolving systems. We inferred all the six systems are evolving in nature with different persistence of entity-connections. The List of Bible translation system has fewer persistent entity-connections, while the Retail market system and IMDB movie genre system has moderate persistent entity-connections, whereas List of Multi-sport events and Hadoop-HDFS has many persistent entity-connections. The positive and negative sentiment system has similar persistence of entity-connections. The two evolving systems of different domains are not comparable based on the SNP metric values because an experiment also depends upon factors like: type of entities and their connections. Second, low SNP metric value does not reflect unstable systems, but it reflects few persistent entity-connections.

**Threat to validity:** The Algorithm SysNet-Analytics uses NRM, thus pros-cons for mining SNERs depends on rule mining. The advantage of rule mining techniques is that it results in a white-box set of associated entities, and their disadvantage is that it generates a large number of rules, which are difficult to be analyzed by humans [28]. However, we optimized the number NERs as SNERs using (stability ≥ minStab).

## VII. RELATED WORKS & DISCUSSIONS

There is intense research activity today involving networks that change over time and concerning such diverse themes as mining: Graph Evolution Rules (GERs) by Berlingerio et al. [18]; Link Formation Rules (LFRs) by Leung et al. [19]; Graph-Pattern Association Rules (GPARs) by Fan et al. [20]; EvoMine tool by Scharwächter et al. [21]; Graph Temporal Association Rules (GTARs) by Namaki et al. [22]; Attribute Evolution Rules (AERs) by Fournier-Viger et al. [23]; and TACOs (TemporAl event prediCtiOn rules) by Wenfei et al. [24]. Böttcher et al. [25] demonstrated association mining to detect changes and retrieved a vast number of change rules. Liu et al. [26] analysed temporal databases to identify trend rules and discarded the unstable rules based on statistical calculation. Shokoohi et al. [27] proposed *time series rules* that are identified using time series motifs. Ale and Rossi [9] presented the temporal rule mining.

Like others [18]-[24], we also retrieved NERs. Assuming NERs as a concept equivalent to rules given in Table III i.e., GERs, LFRs, GPARs etc. First, we introduced a novel method to retrieve NERs, and then optimized the number of SNERs over multiple states. Compare the reduction in the number of rules in the fourth and fifth column of Table II i.e., SNER Count and Total NER Count. This shows a smaller number of SNERs are more interesting and stable than NERs.

TABLE III DIRECT COMPARISON WITH EXISTING STATE-OF-THE-ARTS.

| Rules | Contribution | Future Work |
| --- | --- | --- |
| GERs (2009) | Describes the local changes and applied to four real-world networks. | Investigate rule confidence to predict graph evolution. |
| LFRs (2010) | Edge labels but unlabeled nodes are applied to two real-world datasets. LFR contains link patterns as dyadic and/or triadic structures in social networks | Study of rules at a certain time point and multigraphs. |
| GPARs (2015) | Discover regularities between entities on social media graphs with a parallel (scalable) algorithm. Help in marketing by identifying and influencing customers. | To support graph patterns as consequent and provide other matching semantics as graph simulation. |
| EvoMine (2016) | Mines graphs - edge insertions and deletions - and - node and edge relabelling. Provide results for comparing EvoMine with the GERM and LFR miner. | Time and space analysis for quality and confidence of rules. |
| GTARs (2017) | Class of temporal association rules and used for activity prediction. | Exploring other quality metrics of rules and rules over graphs. |
| AERs (2020) | Demonstrated changes of attributed values of multiple vertices on dynamic graphs. | To discover concise AERs with specific temporal constraints. |
| TACOs (2022) | Demonstrated a system TASTE to discover TACOs on temporal graphs. | To make TASTE for finance and real-time prediction. |
| SNERs (2019 [12] and this paper) | We defined stability of NERs over states to retrieve Stable NERs. We defined persistence metrics. Shown System Evolution Analytics on six state series. | Other mining techniques with network evolution to make new hybrid mining. Application on Big graphs of other systems. |

Usually, network and rule mining are independent of time. Evolution (temporal) mining is independent of inter-connected entities. Our approach is better in the sense that it intelligently combines network, rule, and evolution mining.

*System Evolution Analytics* also includes. The subgraph evolution mining to calculate *System Network Complexity* [29]. Our proposed *Deep Evolution Learning* led to the construction of *System Evolution Recommender* tool [30][31], which further led to the invention of *System Neural Network* [32]. The first author applied the concept of *minStab* to retrieve significant *Stable Hate Rules* over multiple datasets and contexts [33].

## VIII. CONCLUSIONS

We defined evolution rules, stable rules, minStab, persistence metrics, and SNP metric about entity-connections. Our SysNet-Analytics approach is applicable to a set of evolving networks. The approach automatically retrieved evolution and stable rules, which are useful to calculate persistence of entity-connections. We proposed the SysNet-Analytics algorithm for mining NERs and SNERs. The algorithm analyses a pre-evolved system using a threshold minStab to generate SNERs. Using the mining information, we introduced a formula for *persistence metric* and its type *SNP metric*, which quantifies the persistence of entity-connections in an evolving system. We demonstrated real-world applications for four domains based on the six State-Series of six different evolving systems. Our SysNet-Analytics Tool automatically retrieves NERs and SNERs that can help to deduce system evolution information, which is interesting and non-obvious. We calculated nine SNP metric values for nine experiments on each evolving system. The SNP metric measures persistence of entity-connections for an evolving system. In future, other mining techniques with network evolution analytics can produce another hybrid mining approach. Our approach can be applicable to large networks or big graphs. We plan to strengthen our experimentation with other datasets.